# The loss of the property of locality of the kernel in high-dimensional Gaussian process regression on the example of the fitting of molecular potential energy surfaces

Sergei Manzhos[1], Manabu Ihara[2]

School of Materials and Chemical Technology, Tokyo Institute of Technology, Ookayama 2-12-1, Meguro-ku, Tokyo 152-8552 Japan

## Abstract

Kernel based methods including Gaussian process regression (GPR) and generally kernel ridge regression (KRR) have been finding increasing use in computational chemistry, including the fitting of potential energy surfaces and density functionals in high-dimensional feature spaces. Kernels of the Matern family such as Gaussian-like kernels (basis functions) are often used, which allows imparting them the meaning of covariance functions and formulating GPR as an estimator of the mean of a Gaussian distribution. The notion of locality of the kernel is critical for this interpretation. It is also critical to the formulation of multi-zeta type basis functions widely used in computational chemistry We show, on the example of fitting of molecular potential energy surfaces of increasing dimensionality, the practical disappearance of the property of locality of a Gaussian-like kernel in high dimensionality. We also formulate a multi-zeta approach to the kernel and show that it significantly improves the quality of regression in low dimensionality but loses any advantage in high dimensionality, which is attributed to the loss of the property of locality.

## 1 Introduction

Kernel methods such as kernel ridge regression (KRR) [1] and Gaussian process regression (GPR) [2] have been gaining increasing popularity in computational chemistry applications, specifically for the construction of multidimensional potential energy surfaces (PES) [3–14] and functionals for DFT (density functional theory) [15,16] including exchange correlation and kinetic energy functionals [17–22]. Both GPR and KRR have the same form of representing a target function $f(x), x \in R^D$,

$$f(x) = \sum_{n=1}^{M} b_n(x) c_n \equiv \sum_{n=1}^{M} b(x, x_n) c_n = f^* c$$

(1.1)

where $b(x, x')$ is the kernel and $f^*$ is a row vector with elements $f_n^* = b(x, x_n)$. The vector of coefficients $c$ is computed as

$$c = K^{-1} f$$

(1.2)

where $f$ is the vector of all known values of $f(x)$, $f_n = f(x_n)$, $n = 1, \ldots, M$, and

$$K = \begin{pmatrix} b(x_1, x_1) + \delta & \cdots & b(x_1, x_M) \\ \vdots & \ddots & \vdots \\ b(x_M, x_M) & \cdots & b(x_M, x_M) + \delta \end{pmatrix}$$

(1.3)

[1] E-mail: manzhos.s.aa@m.titech.ac.jp
[2] E-mail: mihara@chemeng.titech.ac.jp

The parameter $\delta$ added on the diagonal of $K$ has the meaning of the regularization parameter [1] that is used to improve stability (e.g. against the singularity of $K$) and generalization ability of the model. *Eq. (1.1) is a linear regression with basis functions $b_n(x) = b(x, x_n)$ built from the kernel function.* While typically Eqs. (1.1-1.3) with a square, $M \times M$, matrix $K$, a rectangular formulation is possible [23] in which $K$ is of size $M \times N$, whereby fewer basis functions $N$ than points are used. This allows dispensing with the regularization parameter $\delta$ as the coefficient $c$ are determined in the least squares sense (see below).

In the case of GPR, the kernel is chosen as a covariance function $k(x, x')$ between pairs of data, and the matrix $K$ takes the meaning of a variance-covariance matrix. The predicted value of $f(x)$, Eq. (1.1), has the meaning of a mean of a Gaussian distribution [2]. The kernel $k(x, x')$ is typically one of the Matern family of functions [24],

$$k(x, x') = \sigma^2 \frac{2^{1-\nu}}{\Gamma(\nu)} \left( \sqrt{2\nu} \frac{|x - x'|}{l} \right)^\nu K_\nu \left( \sqrt{2\nu} \frac{|x - x'|}{l} \right)$$

(1.4)

where $\Gamma$ is the gamma function, and $K_\nu$ is the modified Bessel function of the second kind. At different values of $\nu$, this function becomes a squared exponential ($\nu \to \infty$), a simple exponential ($\nu=1/2$) and various other widely used kernels (such as Matern3/2 and Matern5/2 for $\nu = 3/2$ and $5/2$, respectively). The value of $\nu$ is often preset, and the length scale $l$ and prefactor $\sigma^2$ are hyperparameters. In particular, the Gaussian-like squared exponential kernel which is also called the RBF (radial basis function) kernel is by far the most widely used:

$$k(x, x') = \exp\left( -\frac{|x - x'|^2}{2l^2} \right)$$

(1.5)

We omitted any prefactor in Eq. (1.5) as it is taken care of by Eq. (1.2) and is also fully correlated with $\delta$. A key property of the Matern kernels / basis functions such as the RBF kernel is their locality, in that the correlation between data points drops with the distance between them. This picture allows one, for example, to consider local methods based on neglect of correlation between faraway data points. *GPR is thus a particular case of KRR that uses localized kernels that can be viewed as covariance functions.*

Gaussian-like basis functions are also widely used in computational chemistry: in electronic structure, in computational spectroscopy and in quantum dynamics [25–28]. In these applications, the argument of locality is often used to justify the use of Gaussian-like functions. Especially in the electronic structure approaches, Gaussian-like (whether finite or infinite support) basis functions are often used in a multi-zeta setting, whereby several basis functions with Gaussian-like radial components of different width are placed on the same center (such as an atomic position) [27–29]. *The idea of a multi-zeta basis is intrinsically* *connected to the idea of the locality of the basis function.* While this locality is ensured in 3D, which is the relevant dimensionality in many types of ab initio calculations, Gaussian-like basis functions are often used in higher-dimensional calculations, including calculations of vibrational spectra and nuclei dynamics [26,27,30–32].

However, as the dimensionality of the space increases, the Gaussian function loses the property of locality. For example, while about 90% of the quadrature of a Gaussian function is concentrated within one standard deviation from the mean, only 10% of the quadrature of a six-dimensional Gaussian comes from within one standard deviation from the mean. This is one of the phenomena forming the so-called curse of dimensionality [33]. Quadrature is the relevant measure of locality in many applications. It is obviously so in quadrature-based methods for the solution of the electronic or nuclei Schrödinger equation [34,35], but it is also so in collocation approaches [36,37] and whenever one computes rmse (root mean square error) over a set of points (e.g. when fitting a function with GPR), which are related to quadratures with idiosyncratic choices of weights [38]. As the dimensionality increases, therefore, one expects a nivellation of the imputed advantages of locality of Gaussian-like functions, such as the advantage of a multi-zeta basis or advantages stemming from diminishing correlation between data points with distance.

In this work we (i) introduce the idea of a multi-zeta kernel and test its performance when fitting PESs of different dimensionality: 3D ($H_2O$), 6D ($H_2CO$), and 15D ($UF_6$). To demonstrate a practical utility of multi-zeta kernels for calculation of observables, we compute the vibrational spectrum of $H_2CO$. (ii) We show that as the dimensionality of the feature space increases, the property of locality of the kernel disappears, in that the values of all elements of the variance-covariance matrix are not significantly different from 1, and the advantage of a double-zeta kernel, which is very significant in 3D, becomes less significant in 6D and disappears in 6D.

## 2  Methods

We use the rectangularized version of KRR/GPR proposed in Ref. [23]. We use $N$ "training" points which serve as basis centers and $M$ auxiliary points and compute a rectangular version of the matrix $K$, a matrix $B$ of size $M \times N$ with elements $B_{mn} = k(x_m, x_n)$, $n = 1, ..., N$, $m = 1, ..., M$. The coefficients $c$ of Eq. 1 are computed as $c = B^+ f'$, where $f'$ is a vector of target values at the $M$ points and $B^+$ is a Moore-Penrose pseudoinverse [39] of $B$. This allows us to dispose of the noise parameter altogether and focus on the length parameter of the kernel. A total of $M$ points are thus used for training even though there are $N$ basis centers. We use $M = 1.4N$. The exact $M/N$ ratio is not a sensitive parameter, and ratios of this order can be recommended based on the results of Ref. [23]. As the input data are normalized, we use an isotropic Gaussian-like RBF kernel

$$k(x, x'|l) = \exp\left(-\frac{|x - x'|^2}{2l^2}\right)$$

(2.1)

The only hyperparameter $l$ can then be optimized by a simple scan. A double-zeta kernel is realized by augmenting the matrix $B$ with values of $k(x, x')$ computed with different length parameters:

$$B = \begin{pmatrix} k(x_1, x_1|l) & \cdots & k(x_1, x_N|l) & k(x_1, x_1|l_2) & \cdots & k(x_1, x_N|l_2) \\ \vdots & \ddots & \vdots & \vdots & \ddots & \vdots \\ k(x_M, x_1|l) & \cdots & k(x_M, x_N|l) & k(x_M, x_1|l_2) & \cdots & k(x_M, x_N|l_2) \end{pmatrix}$$

(2.2)

Then $c = \begin{pmatrix} c_1 \\ c_2 \end{pmatrix}$ and

$$f(x) = \sum_{n=1}^{N} k(x, x_n|l) c_{1,n} + \sum_{n=1}^{N} k(x, x_n|l_2) c_{2,n}$$

(2.3)

In the following, we use $l_2 = 1.5l$ unless indicated otherwise. To test any advantage of the double-zeta kernel, double-zeta kernel calculations were performed using the same amount of training data, i.e. $N$ and $M$, as the calculations using the conventional RBF (single-zeta) kernel; this is made possible by the use of the pseudoinverse of $B$.

We fitted the potential energy surfaces (PES) of $H_2O$, $H_2CO$ and $UF_6$ molecules as examples of 3D, 6D, and 15D regression problems. For the description of the datasets and information about the applications of these functions, see Ref. [40] for $H_2O$, Ref. [32] for $H_2CO$, and Ref. [41] for $UF_6$. We have a total of 10,000 data points in three dimensions (representing Radau coordinates) for $H_2O$ with values of potential energy ranging 0-20,000 cm$^{-1}$ and sampled from the analytic PES of Ref. [42], 120,000 data points in six dimensions (representing molecular bonds and angles) for $H_2CO$ with values of potential energy ranging 0-17,000 cm$^{-1}$ sampled from the analytic PES of [43], and 54,991 data points in 15 dimensions (representing normal mode coordinates of the 15 modes of vibration) for $UF_6$ with values ranging 0-6,629 cm$^{-1}$ sampled with DFT calculations as described in Ref. [41]. The data are available in the supplementary material of Refs. [44,45]. The distributions of the data can also be found in these references. The inputs were normalized to unit variance before regression.

The vibrational spectrum calculations were done for $H_2CO$ on the underlying analytic PES and on PESs fitted with single- and double-zeta kernels with the rectangular collocation approach of Manzhos and Carrington [4,30,32,36,46,47] with a numeric application of the exact space-fixed cartesian coordinate kinetic energy operator [32]. All 120,000 available points were used as the collocation point set. 40,000 Gaussian basis functions were used. The reader is referred to Refs. [4,30,32] for calculation details.

All calculations are performed in Matlab with home-made codes.

## 3  Results

*3.1  $H_2O$*

The results of fits of the three-dimensional $H_2O$ PES with single- and double-zeta kernels with different length parameters *l* are shown in Figure 1. Here and elsewhere, the root mean square error is plotted for a large test set sufficient to judge the global quality of the potential, which was 5000 points in the case of $H_2O$. Here and elsewhere, we plot rmse from three runs to approximately account for differences in rmse due to different random choices of training points. While not sufficient to collect statistics, the three values are indicative of these differences to a sufficient degree for the purpose of this work. In all cases, we observe a significant improvement in the quality of the regression with the double-zeta kernel, which is achieved without an increase in training data. With only 250 training points (basis centers) a double-zeta kernel achieves a test rmse on the order of 1 cm$^{-1}$ with an optimal length parameter, which is spectroscopically accurate [40]. Figure 1 also shows a correlation plot between the predicted and exact values for a representative fit which is visually indistinguishable from the diagonal with a correlation coefficient of 1.00, which is also the case for all other cases considered here and which are therefore not shown. A single-zeta kernel achieves test rmse on the order of 5 cm$^{-1}$. With 500 and 1000 training points (basis centers), single/double zeta kernels achieve test rmse of about 1.5/0.6 and 0.8/0.3 cm$^{-1}$, respectively. The improvement is therefore more significant with fewer training data.

Figure 2 shows the distributions of the values of the variance-covariance matrix $K$ for these cases for the optimal length parameters that minimize the test set error. We note that (i) the optimal length parameters, of about 3-4, are significantly larger than 1, i.e. larger than the characteristic length scale of the data defined by its variance (which has been normalized to 1); (ii) the values of $K$ are non-negligible for all pairs of points, although they become much smaller than 1, with values for a minority of point pairs as small as 0.2-0.4, depending on the case.

*3.2    $H_2CO$*

The results of the fits of the six-dimensional $H_2CO$ PES with single- and double-zeta kernels with different length parameters are shown in Figure 3, where rmse values at a large test set of 50,000 points are shown for different values of the length parameter $l$. In all cases, the use of a double-zeta kernel allows for a significant improvement of the regression quality without increasing the number of training data. Single/double zeta kernels achieve test rmse of about 35/25, 13/6, and 4.6/1.7 cm$^{-1}$ with 500, 1000, and 2000 training points (basis centers), respectively. The correlation plots between the predicted and exact values for these cases with optimal length parameters are shown in Figure 4; all showing a high quality of regression resulting in a spectroscopically accurate PES [4,48].

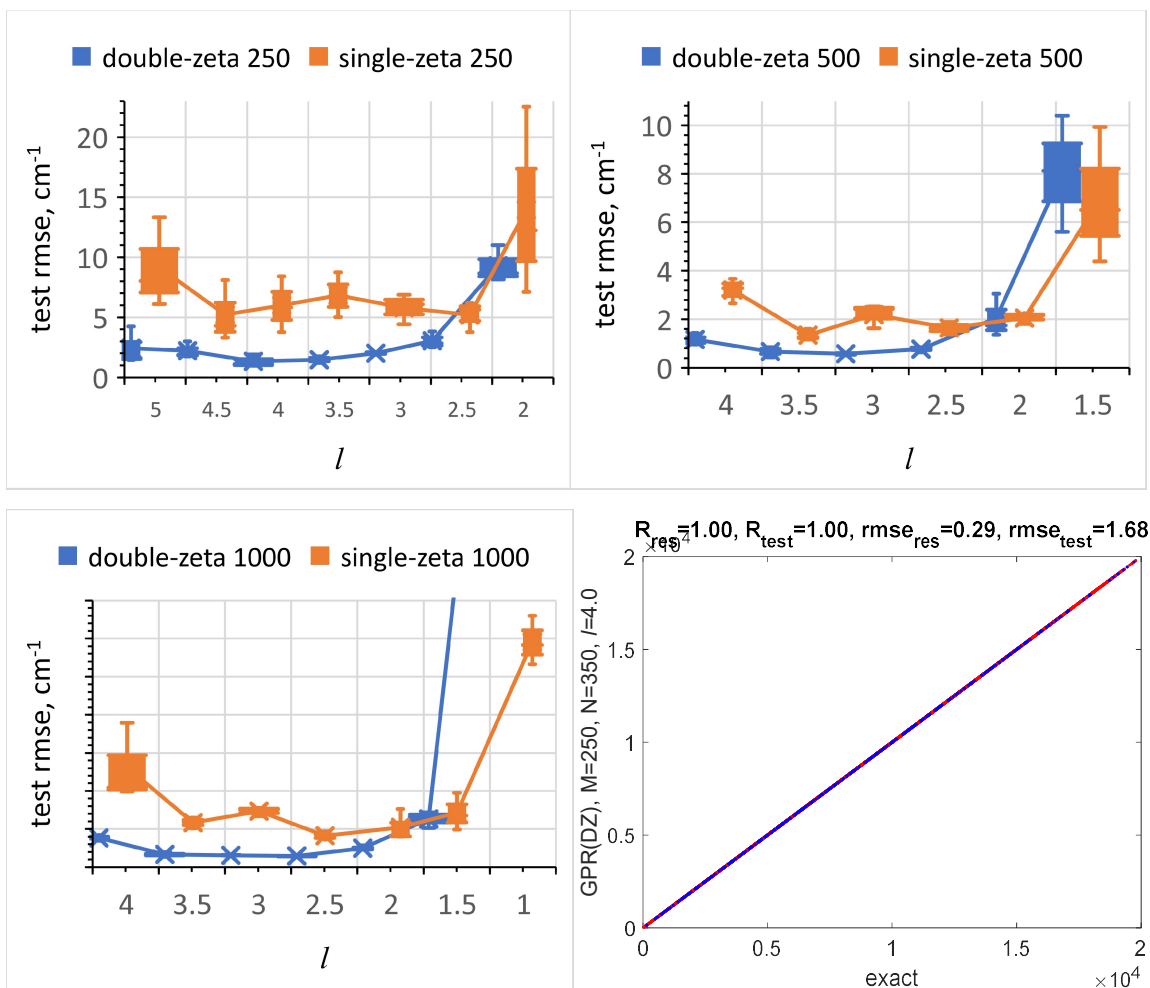

Figure 1. Test set root mean square errors when fitting the potential energy surface of $H_2O$ using 250 (top left), 500 (top right), and 1000 (bottom left) training points (basis centers) with single-zeta and double-zeta kernels with different length parameters $l$. Bottom right: The correlation plot of exact vs predicted energies for an optimal length parameter $l$ when using $N = 250$ basis centers and $M = 350$ in a $M \times N$ rectangular matrix, with a double-zeta kernel. The training points are in blue, and the test points are in red. The correlation coefficient $R$ and rmse values at the test points and $M$ points ("res") are shown on the plot.

.

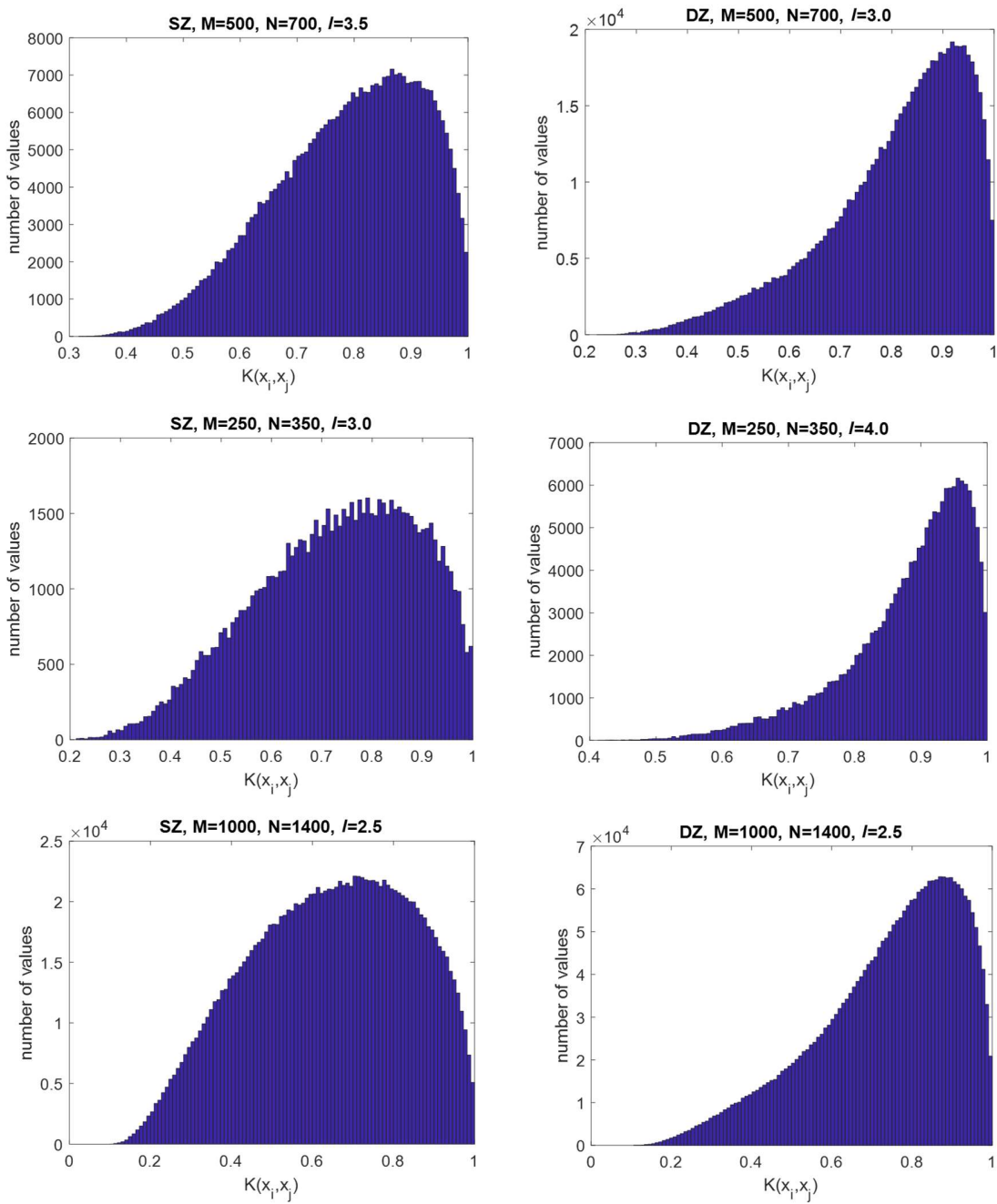

Figure 2. The distributions of the values of the variance covariance matrix $K(x_i, x_j)$ when fitting the potential energy surface of $H_2O$ using different numbers $N$ of basis centers (top row – 250, middle row – 500, bottom row - 1000) and a $M \times N$ rectangular matrix with $M$ values given on the plots, with single-zeta ("SZ", left column) and double-zeta ("DZ", right column) kernels with optimal length parameters $l$ (values given on the plots).

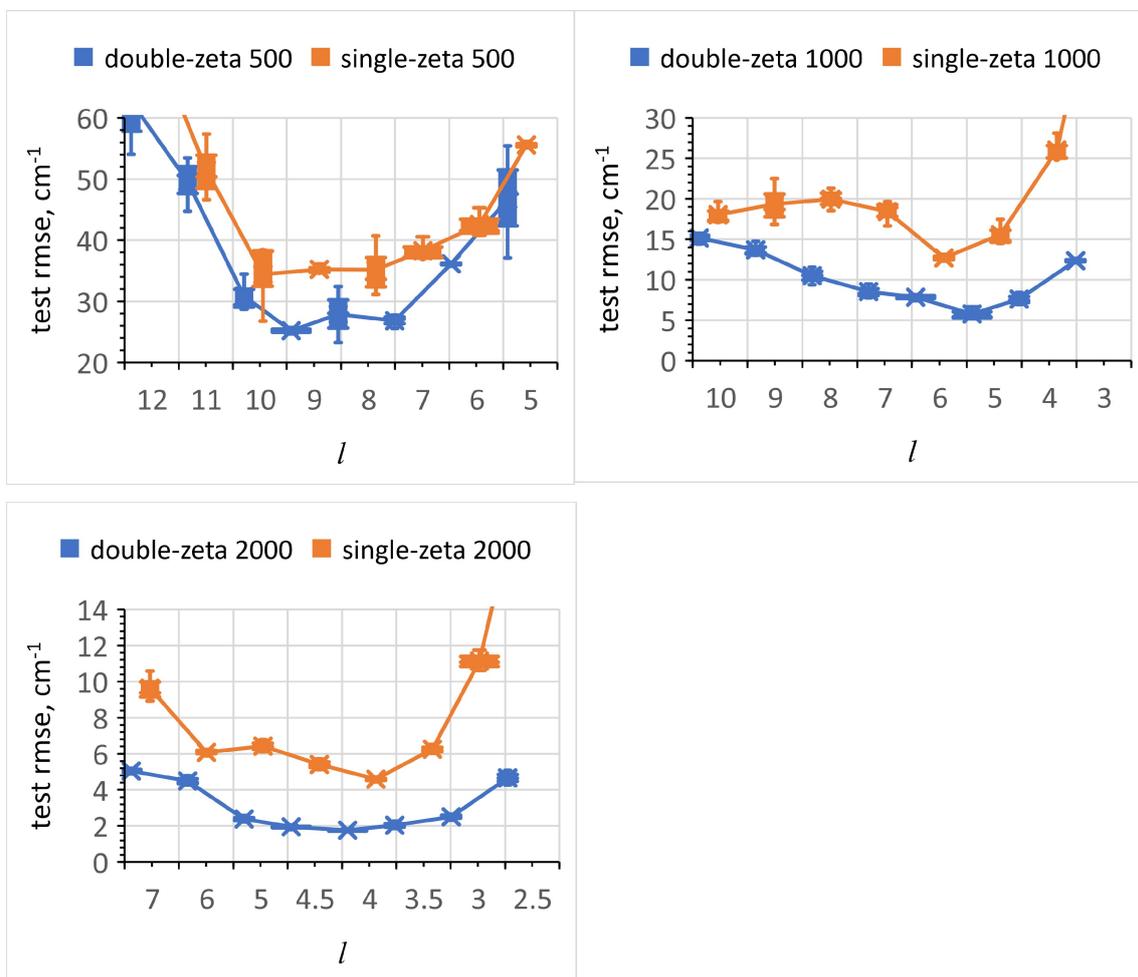

Figure 3. Test set root mean square errors when fitting the potential energy surface of $H_2CO$ using 500 (top left), 1000 (top right), and 2000 (bottom left) training points (basis centers) with single-zeta and double-zeta kernels with different length parameters $l$.

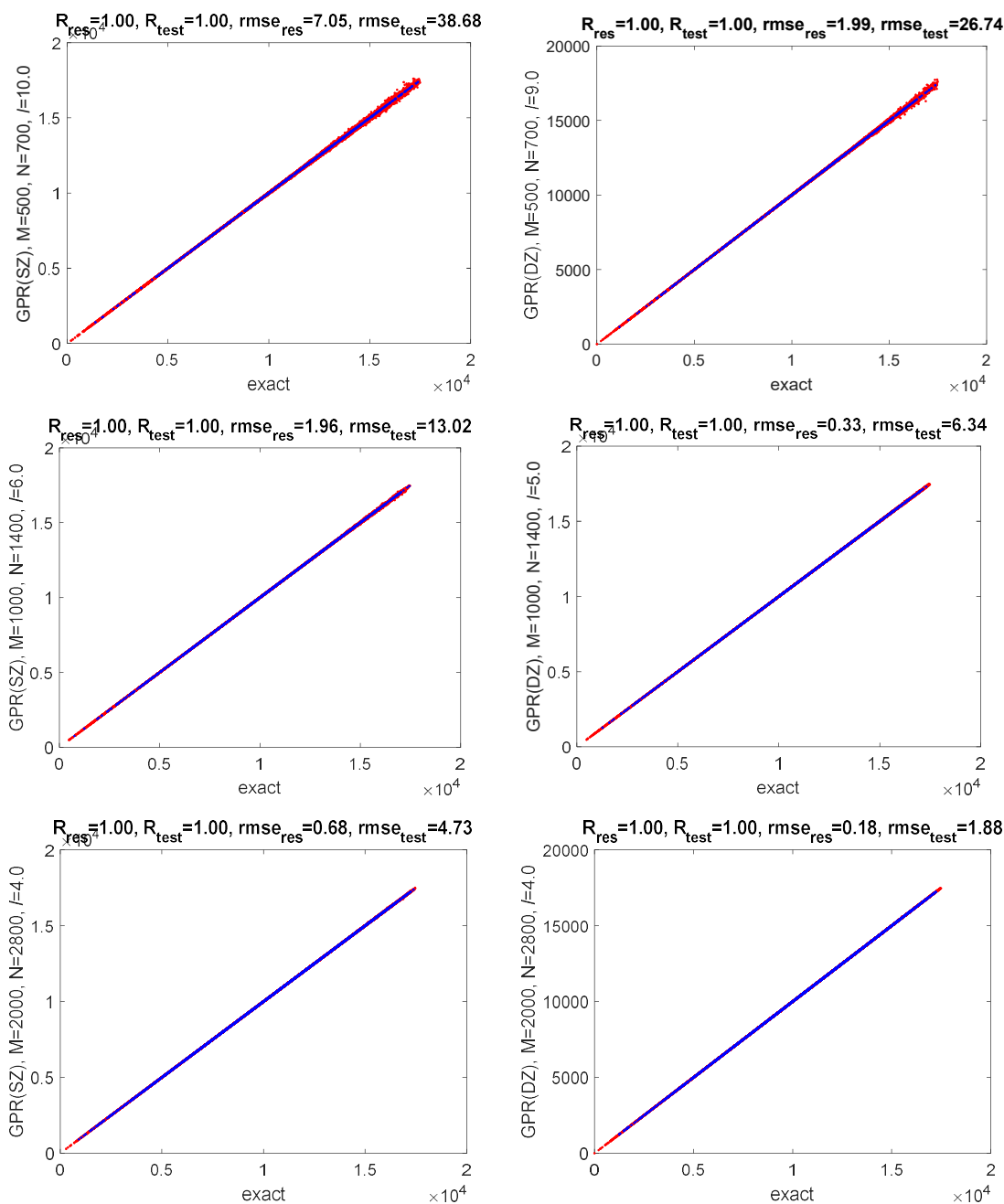

Figure 4. The correlation plots of exact vs predicted H$_2$CO potential energies for optimal length parameters $l$ when using $N$ = 500 (top row), 1000 (middle row), and 2000 (bottom row) training points (basis centers) and a $M \times N$ rectangular matrix, with a single-zeta ("SZ", left column) and double-zeta ("DZ", right column) kernel. The training points are in blue, and the test points are in red. The correlation coefficient $R$ and rmse values at the test points and $M$ points ("res") are shown on the plot.

Figure 5 shows the distributions of the values of the variance-covariance matrix $\mathbf{K}$ for these cases for the optimal length parameters that minimize the test set error. We note that (i) the optimal length parameters, reaching 10 in some cases, are even more significantly larger than 1, i.e. larger than the characteristic length scale of the data, and to a larger extent compared to the 3D case; (ii) the values of $\mathbf{K}$ are non-negligible for all pairs of points. Only a small number of $\mathbf{K}$ entries get as small as 0.3 only when the number of training points is high, while with 500 training points (basis centers) *all entries of $\mathbf{K}$ are on the order of 1*. In this case, the kernel stops being local in a practical sense whereby it is not possible to neglect or even downplay the correlation with far-away data points, which is at the heart of the idea of Matern-type kernels. Note that, contrary to the 3D case, the largest improvement, by almost a factor of 3 in terms of rmse, from the double-zeta kernel vs a single-zeta kernel is now for the case with the most, 2000, training data, which is also the case where the kernel can still be said to be local (due to a lower optimal value of $l$). With 500 training points (basis center), the improvement is much less significant, on the order of 40%.

To confirm the utility of a double-zeta kernel in applications when computing observables, we computed the vibrational spectrum of $H_2CO$ on PESs fitted with single- and double-zeta kernels using $N = 2000$ training points (basis centers) and $M = 2800$. We used the rectangular collocation approach of Manzhos and Carrington using 120,000 collocation points and 40,000 Gaussian basis functions. The results of the comparison of the spectrum quality for the ZPE and the lowest 50 and 100 transition frequencies vs. the calculation done with the same method and parameters on the reference analytic PES are given in Table 1. The spectra are given in the Supporting Information. Both single-zeta and double-zeta PES are spectroscopically accurate, but the PES fitted using a double-zeta kernel to the same amount of training data is much more accurate.

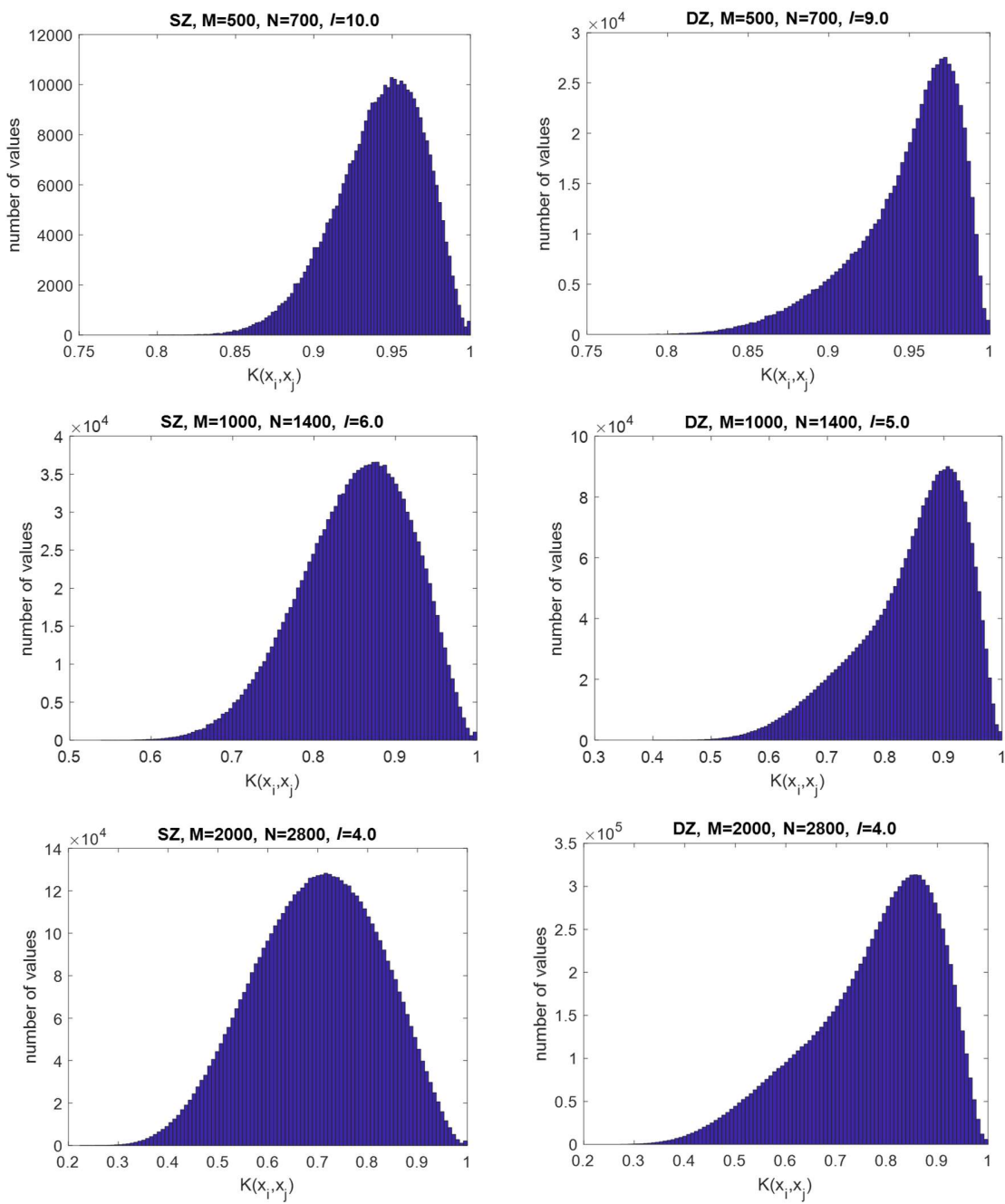

Figure 5. The distributions of the values of the variance covariance matrix $K(\pmb{x_i}, \pmb{x_j})$ when fitting the potential energy surface of $H_2CO$ using different numbers $N$ of basis centers (top row – 500, middle row – 1000, bottom row - 2000) and a $M \times N$ rectangular matrix with $M$ values given on the plots, with single-zeta ("SZ", left column) and double-zeta ("DZ", right column) kernels with optimal length parameters $l$ (values given on the plots).

Table 1. Comparison of spectrum quality – ZPE and mae (mean absolute error) and rmse of the lowest 50 and 100 transition – of spectra computed with rectangular collocation on H$_2$CO PESs fitted with single- and double-zeta kernels using $N = 2000$ training points (basis centers) and $M = 2800$. The errors are vs the calculation done on the reference analytic PES.

| Error in, cm$^{-1}$ | PES | ΔZPE | Lowest 50 transitions | | Lowest 100 transitions | |
|---|---|---|---|---|---|---|
| | | | mae | rmse | mae | rmse |
| Single-zeta kernel | 4.60 | 0.077 | 0.138 | 0.166 | 0.139 | 0.165 |
| Double-zeta kernel | 1.67 | -0.026 | 0.013 | 0.016 | 0.014 | 0.019 |

### 3.3 UF$_6$

The results of the fits of the fifteen-dimensional UF$_6$ PES with single- and double-zeta kernels with different length parameters are shown in Figure 6, where rmse values at a large test set of 40,000 points are shown for different values of the length parameter $l$. The correlation plots between predicted and exact values for these cases with the optimal length parameters are shown in Figure 7. The distributions of the values of entries of the variance variance-covariance matrix $K$ for these cases for the optimal length parameters that minimize the test set error are shown in Figure 8. It is clear from Figure 8 that the optimal-length kernels are not local, with $K$ entries corresponding to even the farthest pairs of points exceeding 0.9 except in one case, where most values are still in excess of 0.75. To this corresponds the absence of any improvement achieved with a double-zeta kernel in this fifteen-dimensional example. The sharp deterioration of the rmse curves computed with a double-zeta kernel at low $l$ is due to an instability which is also a reflection of the loss of the property of locality, whereby a 50% difference in the length parameter ($l_2 = 1.5l$) dies not sufficiently distinguish between the zeta components in 15D. To illustrate this, we also show in Figure 6 the rmse with a double-zeta kernel computed with $l_2 = 5l$, in which case the deterioration is alleviated but the best (with respect to $l$) test set rmse is still not improved vs the single-zeta kernel.

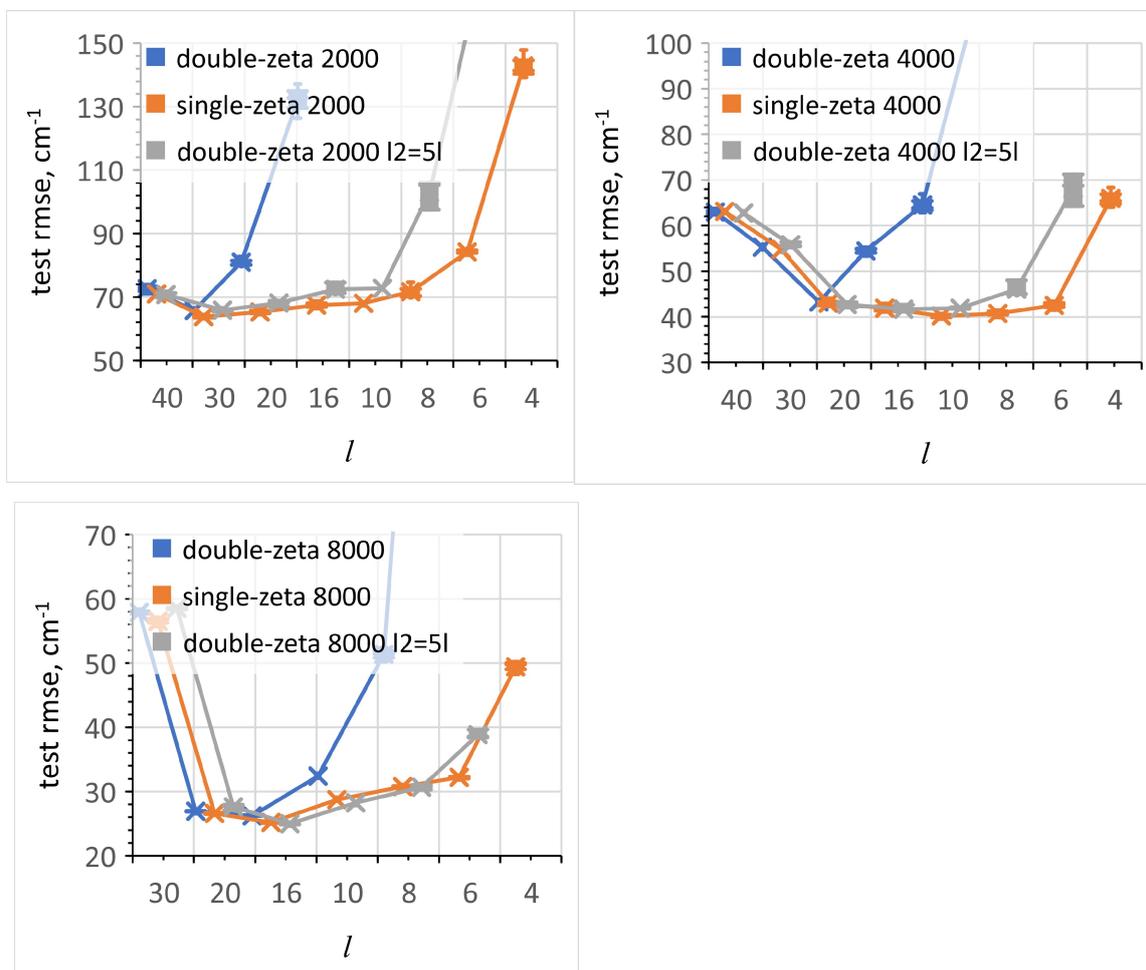

Figure 6. Test set root mean square errors when fitting the potential energy surface of UF$_6$ using 2000 (top left), 4000 (top right), and 8000 (bottom left) training points (basis centers) with single-zeta and double-zeta kernels with different length parameters $l$.

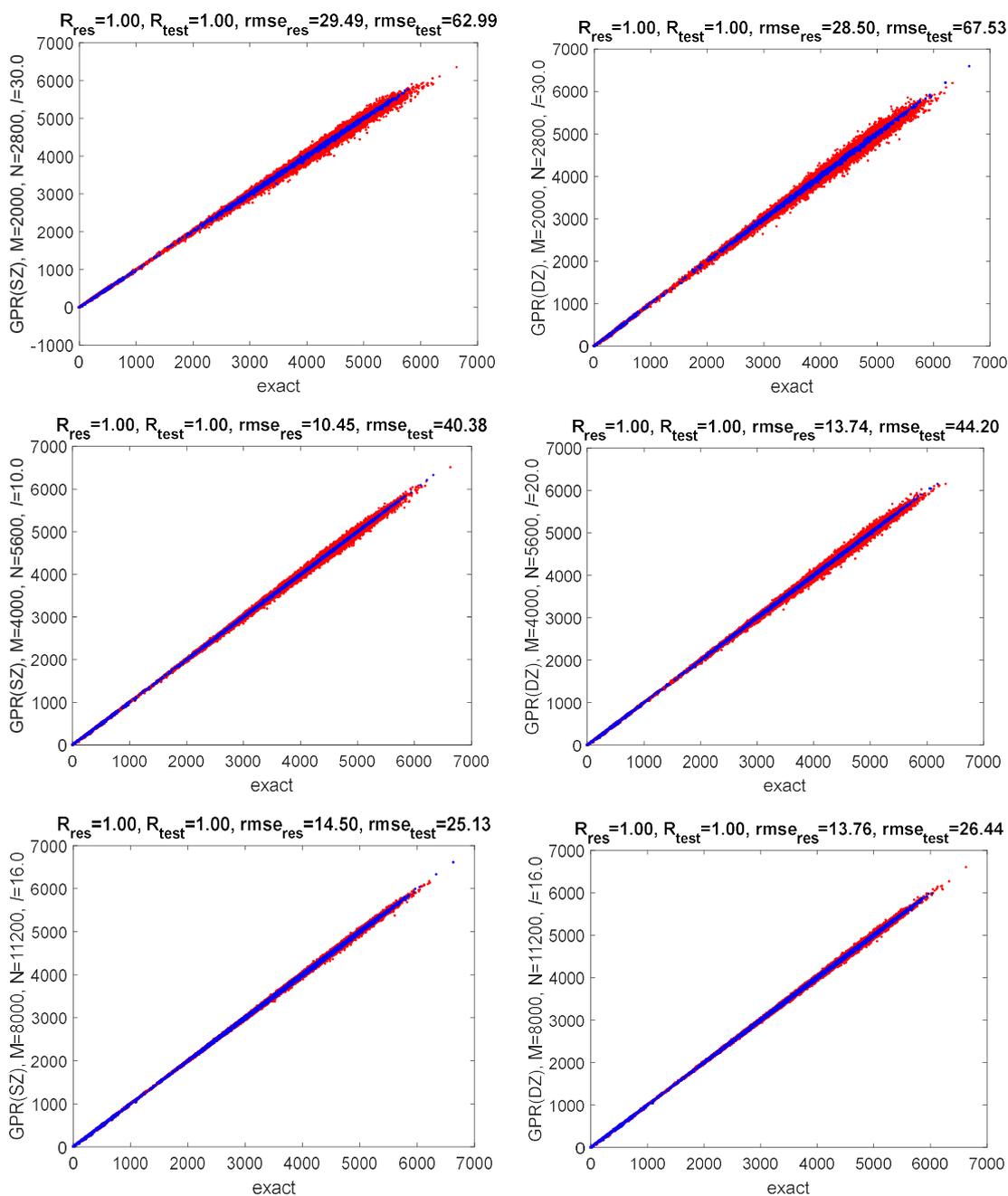

Figure 7. The correlation plots of exact vs predicted UF$_6$ potential energies for optimal length parameters $l$ when using $N = 2000$ (top row), 4000 (middle row), and 8000 (bottom row) training points (basis centers) and a $M \times N$ rectangular matrix, with a single-zeta ("SZ", left column) and double-zeta ("DZ", right column) kernel. The training points are in blue, and the test points are in red. The correlation coefficient $R$ and rmse values at the test points and $M$ points ("res") are shown on the plot

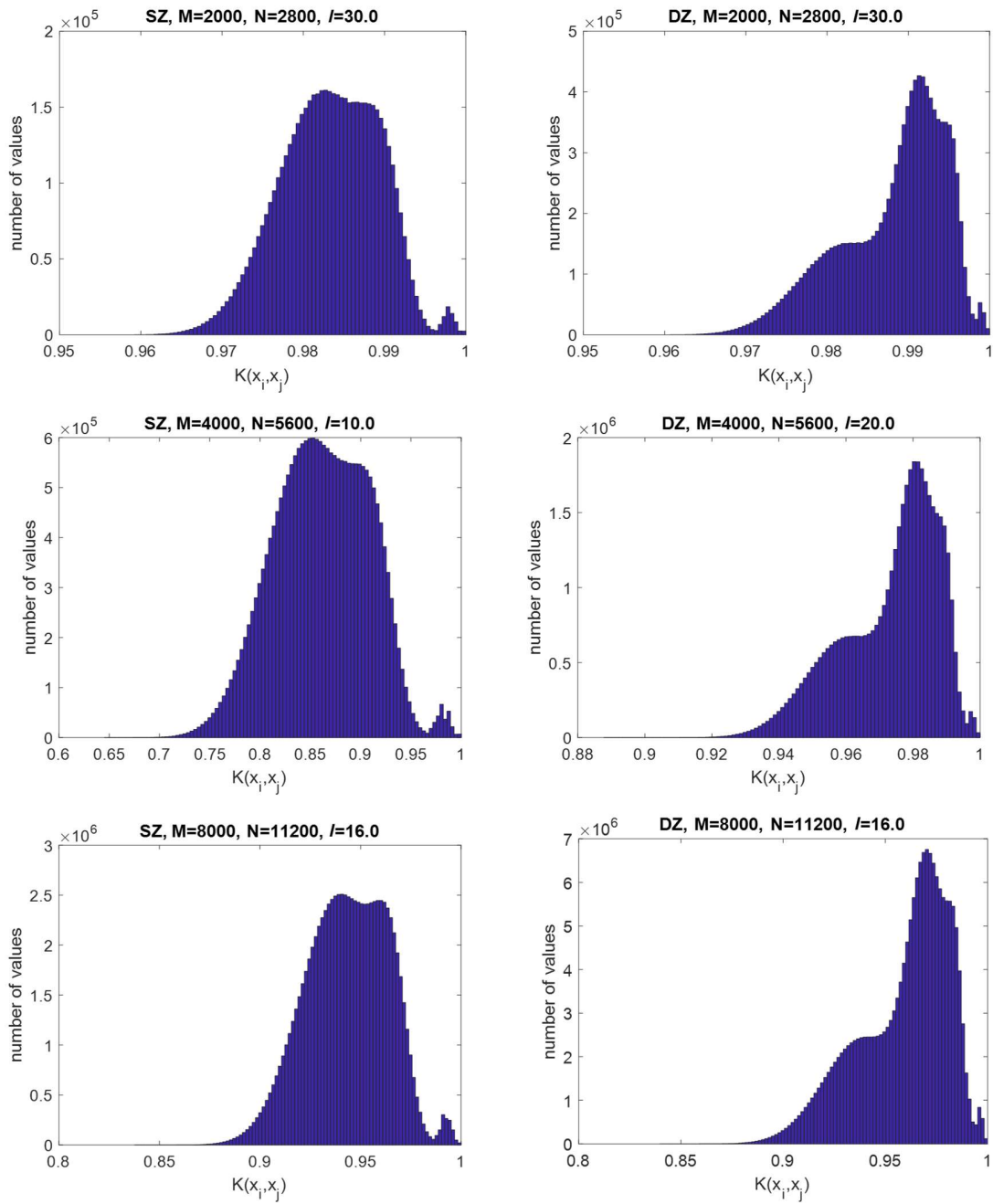

Figure 8. The distributions of the values of the variance covariance matrix $K(\pmb{x}_i,\pmb{x}_j)$ when fitting the potential energy surface of $UF_6$ using different numbers $N$ of basis centers (top row – 2000, middle row – 4000, bottom row - 8000) and a $M{\times}N$ rectangular matrix with $M$ values given on the plots, with single-zeta ("SZ", left column) and double-zeta ("DZ", right column) kernels with optimal length parameters $l$ (values given on the plots).

# 4 Conclusions

We performed kernel ridge regressions with Gaussian-like kernels of potential energy surfaces in three-, six, and fifteen-dimensional configuration spaces ($H_2O$, $H_2CO$ and $UF_6$, respectively). We used a rectangularized version of KRR/GPR that facilitates the selection of hyperparameters [23]. We showed that as the dimensionality of the feature space increases, the property of locality of the kernel disappears (when optimal length parameters are used), in that the values of all elements of the variance-covariance matrix are not significantly different from 1.

We introduced the idea of multi-zeta type kernels and tested it by applying double-zeta kernels to the regression of these PESs. We showed that a double-zeta kernel can significantly improve the quality of regression without increasing the amount of training data when the dimensionality is low enough. A significant improvement was achieved in 3D and less significant in 6D. We also computed the vibrational spectrum of $H_2CO$ to demonstrate the utility of the multi-zeta kernel idea in applications. In 15D, on the contrary, where the kernel loses the property of locality, the advantage of a double-zeta kernel completely disappears, highlighting that the multi-zeta idea is tightly connected to the locality of the basis function.

These results indicate that any deemed advantages of the locality of Gaussian-like basis functions and kernels may be illusory in high dimensionality, and it may be better to use alternative functions. Alternatively, one could use HDMR (high-dimensional model representation) [49–51] -type representations with component functions represented by GPR [5,10,44,45], whereby component functions of sufficiently low dimensionality can preserve the advantages of locality of a Gaussian-like kernel.

# 5 Acknowledgements

S.M. thanks Prof. Tucker Carrington for discussions.# 6 References

# Supporting Information

# The loss of the property of locality of the kernel in high-dimensional Gaussian process regression on the example of the fitting of molecular potential energy surfaces

Sergei Manzhos, Manabu Ihara

**Table S1.** Vibrational spectrum of H2CO computed on the reference analytic PES ("ref") and on PESs fitted with single- and double-zeta kernels. $N = 2000$, $M = 2800$, $l = 4$.

| ref | single-zeta | double-zeta |
|---|---|---|
| ZPE | ZPE | ZPE |
| 5775.06 | 5775.137 | 5775.034 |
| frequency | frequency | frequency |
| 1166.907 | 1166.967 | 1166.904 |
| 1250.604 | 1250.725 | 1250.617 |
| 1500.323 | 1500.132 | 1500.343 |
| 1746.842 | 1746.575 | 1746.848 |
| 2326.962 | 2326.894 | 2326.949 |
| 2421.902 | 2422 | 2421.902 |
| 2498.257 | 2498.337 | 2498.261 |
| 2665.864 | 2665.797 | 2665.872 |
| 2719.951 | 2719.925 | 2719.956 |
| 2780.972 | 2780.803 | 2780.988 |
| 2842.309 | 2842.29 | 2842.311 |
| 2906.051 | 2905.966 | 2906.066 |
| 2999.551 | 2999.286 | 2999.571 |
| 3001.025 | 3000.903 | 3001.026 |
| 3238.966 | 3238.71 | 3238.969 |
| 3472.438 | 3472.127 | 3472.411 |
| 3480.543 | 3480.328 | 3480.521 |
| 3585.834 | 3585.757 | 3585.811 |
| 3675.15 | 3675.176 | 3675.146 |
| 3742.174 | 3742.181 | 3742.16 |
| 3824.549 | 3824.402 | 3824.54 |
| 3886.848 | 3886.872 | 3886.844 |
| 3937.509 | 3937.505 | 3937.513 |
| 3940.169 | 3940.052 | 3940.163 |
| 3995.92 | 3995.957 | 3995.924 |
| 4033.033 | 4032.928 | 4033.052 |
| 4058.185 | 4058.057 | 4058.197 |
| 4085.784 | 4085.778 | 4085.804 |
| 4164.794 | 4164.721 | 4164.796 |
| 4164.794 | 4164.721 | 4164.796 |
| 4195.148 | 4195.04 | 4195.151 |
| 4250.323 | 4250.214 | 4250.312 |
| 4253.174 | 4252.906 | 4253.191 |
| 4336.247 | 4336.14 | 4336.267 |
| 4396.949 | 4396.886 | 4396.947 |
| 4467.968 | 4467.792 | 4467.961 |
| 4501.622 | 4501.385 | 4501.637 |
| 4527.831 | 4527.441 | 4527.849 |
| 4572.707 | 4572.495 | 4572.726 |
| 4624.524 | 4624.384 | 4624.523 |
| 4628.436 | 4628.134 | 4628.412 |
| 4729.214 | 4729.017 | 4729.206 |
| 4734.107 | 4733.884 | 4734.08 |
| 4744.58 | 4744.361 | 4744.54 |
| 4843.013 | 4842.857 | 4842.988 |
| 4926.106 | 4926.048 | 4926.091 |
| 4953.53 | 4953.251 | 4953.497 |
| 4975.431 | 4975.185 | 4975.41 |
| 4983.337 | 4983.284 | 4983.306 |
| 5043.969 | 5043.864 | 5043.947 |
| 5091.539 | 5091.354 | 5091.507 |
| 5109.228 | 5109.224 | 5109.224 |
| 5142.02 | 5141.948 | 5142.015 |
| 5154.484 | 5154.463 | 5154.483 |
| 5171.453 | 5171.151 | 5171.397 |
| 5195.089 | 5195.002 | 5195.085 |
| 5204.563 | 5204.359 | 5204.566 |
| 5244.971 | 5244.994 | 5244.976 |
| 5274.972 | 5274.889 | 5274.983 |
| 5317.084 | 5316.974 | 5317.081 |
| 5320.914 | 5320.777 | 5320.901 |
| 5327.128 | 5327.057 | 5327.129 |
| 5361.437 | 5361.404 | 5361.431 |
| 5391.75 | 5391.691 | 5391.759 |
| 5409.95 | 5409.781 | 5409.947 |
| 5418.631 | 5418.555 | 5418.62 |
| 5434.314 | 5434.238 | 5434.32 |
| 5460.96 | 5460.864 | 5461.029 |
| 5489.718 | 5489.721 | 5489.719 |
| 5493.608 | 5493.503 | 5493.59 |
| 5531.04 | 5530.896 | 5531.077 |
| 5544.122 | 5544.021 | 5544.129 |
| 5552.489 | 5552.391 | 5552.488 |
| 5627.777 | 5627.726 | 5627.767 |
| 5651.167 | 5651.025 | 5651.169 |
| 5667.459 | 5667.396 | 5667.453 |
| 5674.457 | 5674.29 | 5674.459 |
| 5674.457 | 5674.29 | 5674.459 |
| 5688.398 | 5688.269 | 5688.394 |
| 5717.379 | 5717.31 | 5717.39 |
| 5727.399 | 5727.14 | 5727.413 |
| 5768.252 | 5768.087 | 5768.263 |
| 5769.741 | 5769.422 | 5769.719 |
| 5775.087 | 5774.813 | 5775.115 |
| 5812.134 | 5811.955 | 5812.141 |
| 5824.316 | 5824.208 | 5824.332 |
| 5887.555 | 5887.504 | 5887.518 |
| 5888.895 | 5888.819 | 5888.891 |
| 5897.774 | 5897.485 | 5897.723 |
| 5937.479 | 5937.323 | 5937.466 |
| 5986.617 | 5986.388 | 5986.605 |
| 5986.617 | 5986.388 | 5986.605 |
| 5996.164 | 5995.913 | 5996.162 |
| 6004.915 | 6004.653 | 6004.875 |
| 6052.611 | 6052.412 | 6052.612 |
| 6095.789 | 6095.558 | 6095.757 |
| 6103.235 | 6103.124 | 6103.216 |
| 6118.045 | 6117.742 | 6118.022 |
| 6176.249 | 6176.125 | 6176.221 |
| $\Delta$ZPE | 0.077049 | -0.026245 |
| mae 50 levels | 0.137795 | 0.013264 |
| rmse 50 levels | 0.166402 | 0.016411 |
| mae 100 levels | 0.138902 | 0.013897 |
| rmse 100 levels | 0.165028 | 0.01894 |